\documentclass[conference]{IEEEtran}
\usepackage{graphicx}
\usepackage{parskip}
\usepackage[labelfont=bf,textfont=it]{caption}
\usepackage{url}
\usepackage[latin1]{inputenc}
\usepackage[T1]{fontenc}
\usepackage{type1ec}
\usepackage{color, colortbl}
\usepackage{hyphenat}
\usepackage{tikz}
\usepackage{filecontents}
\usepackage{graphicx}
\usepackage{pgfplots}
\usepackage{amsmath}
\usepackage{mathtools}
\usepackage{multirow,bigdelim}
\usepackage{array}
\usepackage{setspace}

\inputencoding{utf8}
\definecolor{Vermelho}{RGB}{203,85,85}
\definecolor{Azul}{RGB}{149,179,215}
\definecolor{Verde}{RGB}{101,159,101}
\definecolor{Roxo}{RGB}{169,150,192}
\definecolor{Amarelo}{RGB}{238,221,92}
\definecolor{Rosa}{RGB}{245,159,220}
\definecolor{DarkGray}{gray}{0.7}
\definecolor{Gray}{gray}{0.8}
\definecolor{LightGray}{gray}{0.9}

\newcommand{\medio}{\cellcolor{Gray}}

\DeclareMathSizes{5}{5}{5}{5}

\begin{document}
    \title{Setting Players' Behaviors in World of Warcraft\\ through Semi-Supervised Learning}

    \author{
        \IEEEauthorblockN{Marcelo Souza Nery}
        \IEEEauthorblockA{Pontifícia Universidade Católica de Minas Gerais\\Instituto de Ciências Exatas e Informática\\Departamento de Ciência da Computação\\
        Belo Horizonte, MG - Brasil \\msnery@gmail.com}
        \\
        \IEEEauthorblockN{Roque Anderson S. Teixeira}
        \IEEEauthorblockA{Universidade Federal de Minas Gerais \\ Instituto de Ciências Exatas\\ Departamento de Ciência da Computação\\ Belo Horizonte, MG - Brasil\\ roqueandersonst@ufmg.br}
        \and
        \IEEEauthorblockN{Victor do Nascimento Silva}
        \IEEEauthorblockA{Pontifícia Universidade Católica de Minas Gerais\\ Instituto de Ciências Exatas e Informática\\ Departamento de Ciência da Computação\\Belo Horizonte, MG - Brasil\\vnsilva@outlook.com}
        \\
        \IEEEauthorblockN{Adriano Alonso Veloso}
        \IEEEauthorblockA{Universidade Federal de Minas Gerais\\ Instituto de Ciências Exatas\\Departamento de Ciência da Computação\\Belo Horizonte, MG - Brasil\\adrianov@dcc.ufmg.br}
    }

    \maketitle

\begin{abstract}

Digital games are one of the major and most important fields on the entertainment domain, which also involves cinema and music. Numerous attempts have been done to improve the quality of the games including more realistic artistic production and computer science. Assessing the player's behavior, a task known as player modeling, is currently the need of the hour which leads to possible improvements in terms of: (i) better game interaction experience, (ii) better exploitation of the relationship between players, and (iii) increasing/maintaining the number of players interested in the game. In this paper we model players using the basic four behaviors proposed in \cite{BartleArtigo}, namely: achiever, explorer, socializer and killer. Our analysis is carried out using data obtained from the game ``World of Warcraft'' over 3 years (2006 $-$ 2009). We employ a semi-supervised learning technique in order to find out characteristics that possibly impact player's behavior.

\end{abstract}


\section{Introduction}

The growing ability to assess real-time actions associated with interactions between players and intelligent agents in digital games, makes them attractive to explore, investigate, develop and test different computational techniques from the Artificial Intelligence (AI) techniques to classic human behavior \cite{MIHALY1990}, and statistics, such as Machine Learning (ML). In the last years, digital games have been used for such a task \cite{LAIRD2001}, \cite{ADOBBATI2001}. Furthermore, \cite{RussellNorvig2010} adds that, for researchers in computing, the abstract nature of games makes them objects of appeal for studying algorithms, and the players are usually restricted to a small set of possible actions, defined by a group small and precise rules. Finally, the quality of a digital game is directly related to their entertainment value, and this, among other factors, with the interactivity with the game \cite{TOZOUR2002b}, \cite{SCOTT2002}.

However, it is hard to obtain datasets related to popular games, since the companies owning their rights usually do not provide such data for research (i.e., this data is often used to perform their marketing strategies). The work of data gathering performed by \cite{LEE2011} allows researchers to explore the universe of the game ``World of Warcraft'', and to extract useful information for analysis of the game and its players. This paper makes an analysis of this game, identifying the player behaviors following the taxonomy proposed by Bartle in \cite{BartleArtigo} and \cite{BARTLE2004}, and evaluates the impact of the relationship between the behaviors of these players.

\begin{figure}[!t]
\begin{center}
\includegraphics[scale=0.45]{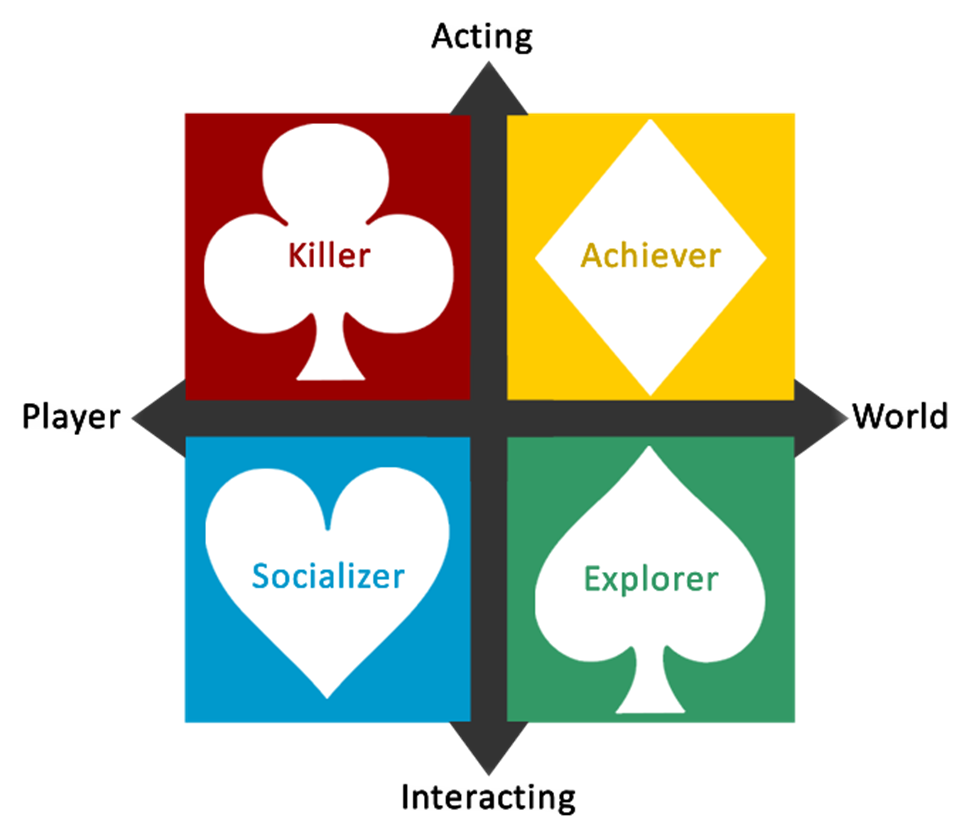}
\caption{The behaviors identified by Bartle: \emph{achiever}, \emph{explorer}, \emph{socializer} and \emph{killer}. The axis \emph{player-world} and \emph{acting-interacting} define how the players act and interact in the game or with/against other players, thus creating these four behaviors. The suit cards' symbols are references used by Bartle to represent the main characteristics of the behaviors.}
\label{fig:Bartle1}
\end{center}
\end{figure}

\section{Related Works}

In recent years, the market for digital games has seen a large growth driven by more elaborate graphics and sound effects, that is, the gaming industry has focused its efforts on developing more powerful graphics cards and advanced audio. However, the quality of a digital game is directly related to their entertainment value and this, in turn, with the interactivity of the game, which can occur through automated intelligent computational agents or non-playable character (NPC), as shown in \cite{TOZOUR2002b, SCOTT2002}, or by the interaction between the players and the game \cite{RABIN2010, KOSTER2005, salen2004rules, MIHALY1990}.

One of the current research areas of interest in digital games is to determine the style or behavior of a player: how she/he behaves while playing, and what are her/his reactions to certain types of game rules. This task is known as \emph{player modeling} and it helps game developers or game designers to create experiences with appeals specifically devoted to each behavior.

Recent research attempts have tried to expand and establish new player behaviors, although there are some behaviors already established in the literature \cite{Keirsey, LANKVELD2009, MACHADO2011, SMITH2011, SMITH2011b, STEWART2013}. However, the pioneering work of \cite{BartleArtigo}, which introduced approaches to create such behaviors is so far the most cited and used. Bartle identified four basic behaviors of players:

\begin{itemize}
\item \emph{Achievers}: prefer to earn points, to evolve the character, to acquire equipment and other concrete measures of success in the game;

\item \emph{Explorers}: prefer to know the whole game environment, to discover secret areas, to find out 3-D modeling errors or programming errors, easter eggs and know all the possible items, such as monsters and maps;

\item \emph{Socializers}: prefer to play online for social pleasure, to interact with other players and to naturally evolve the character;

\item \emph{Killers}: prefer to develop the character, but not with the intention of obtaining merits as achievers. In constrast, they are interested in competing against other players or against NPC enemies that are more powerful and complex.
\end{itemize}

\begin{figure}[!t]
\begin{center}
\includegraphics[scale=0.35]{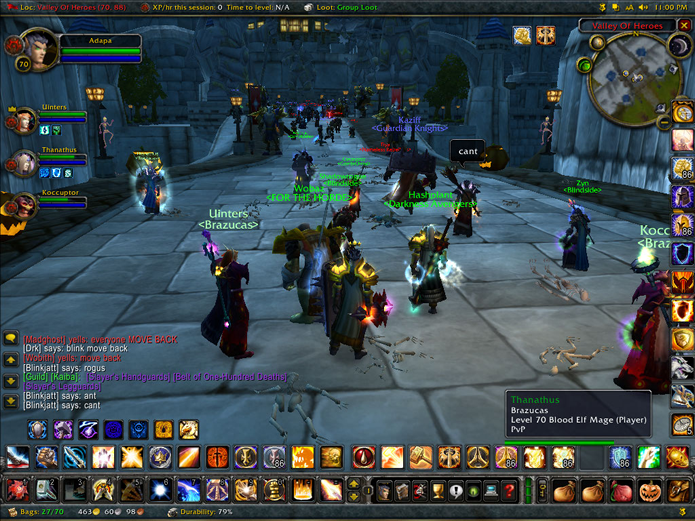}
\caption{Screenshot of the game \emph{World of Warcraft} showing the interface (as spells, minimap, player's status and so on) and various players during a game's session.}
\label{fig:WoW}
\end{center}
\end{figure}

Figure \ref{fig:Bartle1} shows the Bartle graph for the four behaviors, axles called \emph{player-world} and \emph{acting-interacting}, just illustrating the interests of each player behavior: achievers act in the world, ie, extract resources from the system/game for own good; explorers interact with the world, discovering its limits and using your knowledge to; Socializers interact with other players and killers act with other players.

\begin{figure}[!t]
\begin{center}
\includegraphics[scale=0.4]{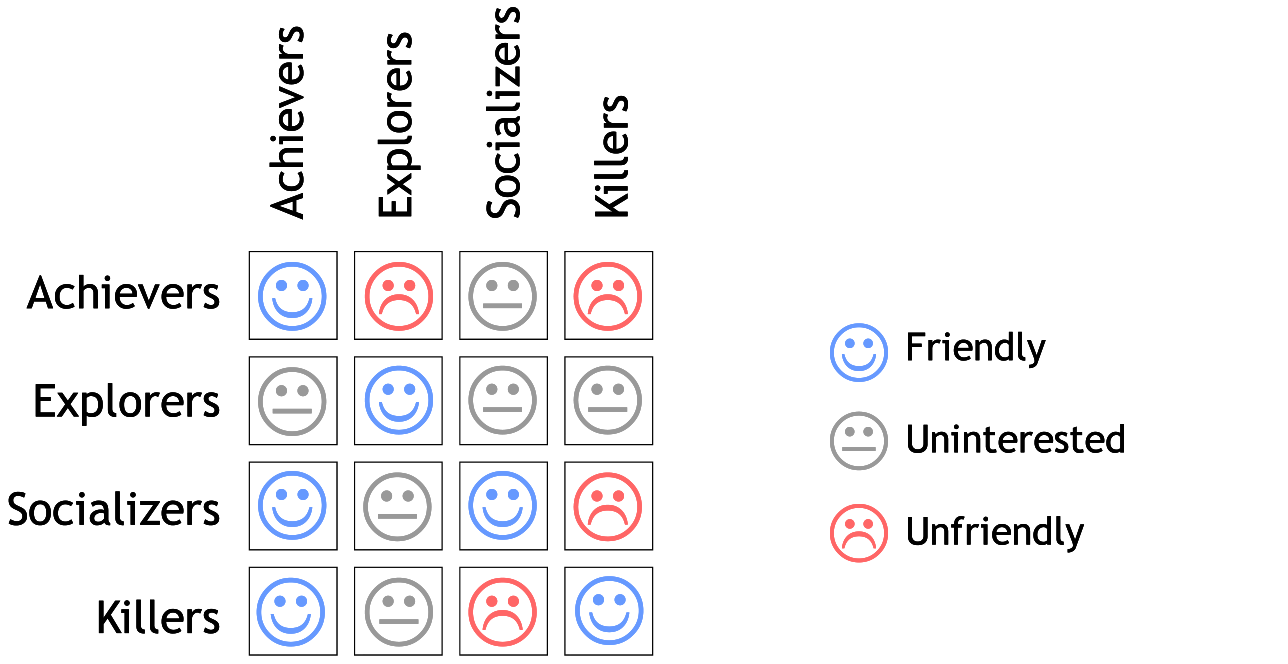}
\caption{The synthesized relationship between each behavior as observed by Bartle --- read this relationship table first horizontally and then vertically. For example, the killer behavior does not relate well to the socializer, and this indicates that the presence of many types of this player inhibits the existence of the other, and for this reason such table is not symmetrical}
\label{fig:Bartle2}
\end{center}
\end{figure}

Figure \ref{fig:Bartle2} synthetizes the relationship between each behavior as observed by Bartle --- reading this relationship table is made first horizontally and then vertically. For example, the player killer does not relate well to the player socializer, and this indicates that the presence of many types player that inhibits the existence of the other, for this reason such a table is not symmetrical. In his article, Bartle describes in detail the reasons for such relationships, as well as features for the game designer what to include/exclude in game to favor the growth/extinction of each of these profiles. Moreover, Bartle also has a feature that facilitates identifying each profile: the players' way to express verbally by gaming chats. While, for example, socializers talk about his personal life, talk a lot and use many emoticons, the killers almost do not talk and maintains a dialogue monosyllabic.

Besides the data analysis done on old online games, or MUD (multi-user dungeon), was also developed a survey called \emph{Bartle Test} \cite{BARTLETest} where players can try to identify themselves in these behaviors. In the field of ML, many works in the area of digital games have been made with different goals, how to predict the players quit a game in \cite{TARNG2009}, identifying robots (or \emph{bots}) playing instead of players in \cite{CHEN2008, THAWONMAS2008, PAO2010} use of SVM to identify players' patterns in \cite{Burges98atutorial} and others.

\subsection{World of Warcraft}

MMO, acronym for Massively Multiplayer Online, has revolutionize major digital games in the world, specially by increasing the number and diversity of players
\cite{Williams06fromtree}. These games bring together thousands of players in an environment where they can easily interact with each other. There are several sub-classifications of MMO. Particularly,  the MMORPG, or Massively Multiplayer Online Rolling Playing Game, is the most widespread one. In this case, there is a persistent and parallel virtual world where players are represented by customizable avatars. The game presented in this work, the World of Warcraft (WoW) as shown in Figure \ref{fig:WoW}, is considered the most played MMORPG in the world, with around 8.3 million active players.

WoW was released in November 2004 as a continuation of the series of Warcraft games, from the company Blizzard Entertainment. Since its launch, the game has already received several patches of expansion, including \emph{The Burning Crusade} (released January 2007) and \emph{Wrath of the Lich King} (released November 2013). The game is divided into two factions, the Horde and the Alliance. Each faction is divided into 5 races and 10 general classes, plus 229 zones or regions. The data collected for this study comprise only Horde players.

The most important changes produced by the patches are associated with \emph{The Burning Crusade}, which increases the maximum level (60 to 70) and added two new races (Blood Elves and Draenei), a new land called Outland, with new maps, monsters, dungeons and cities. In addition to these changes, the patch allowed Horde players to have access to the Paladin class which was previously exclusive to the Alliance faction. A new combat map (Players versus Players or PvP), Eye of the Storm, was also made available. The second patch released during our data gathering process, \emph{Wrath of the Lich King}, further increased the maximum level to 80, and added a new race (Death Knight, for users who already possess some character with a level greater than 55) and a new land called Northrend.

The annual egress rate of WoW is approximately 1.3 million players per year, but the growing popularity of the game is proportional to this egress rate \cite{CNET2013}, so the active population is considered stable.

\section{Methodology and Development}

In this study we used the dataset collected in \cite{LEE2011}. The original dataset has 12 attributes: Date/Time, User ID (protecting player privacy), Guild (or group to which the player belongs), Level, Race, Class, Zone Name Visited, and three dummy attributes. Data was collected during 1,107 days every 10 minutes (from January 2006 to January 2009), before the release of the \emph{Cataclysm} patch (i.e., the maximum possible evolution in the game was 80).

We first inspected the data for better understanding it. Then, we produced information with better potential for classification than the original data, with particular interest in obtaining useful information to assist in the classification of the four categories associated with Bartle's behaviors. Thus, the attributes ``user ID'', ``Race'' and ``Class'' were kept and the attribute ``Date/Time'' was transformed into total playing time during the period analyzed. The attribute ``Guild'' was maintained. Further, it was added a new  attribute, ``Number of guilds'', that counts how many guilds the player joined during the period analyzed.

The attribute ``Zone Name'' was considered whether it would be a city or a PvP area, and classify it by its level. The interest in classifying the areas such as city or a PVP area is that these are regions where some special behaviors prefer to spend most of their time in the game (as is the case killers and socializers). For the zone level, we used the information from the game manual, which are classified according to the average level appropriate for each player. Thus, it was divided into zones as neutral levels, beginner, low, medium and high. Finally, the original attribute ``Level'' was transformed into four attributes: ``Initial Level'', ``Final Level'', ``Evolution'' (which is the final level minus the initial) and ``Speed of evolution'' (which relates to the evolution with time spent during the period analyzed).

\begin{figure}[!t]
\begin{center}
\includegraphics[scale=0.8]{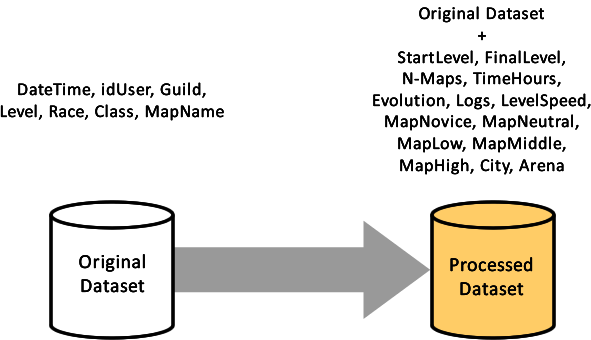}
\caption{(Left) Original dataset. (Right) Processed dataset.}
\label{fig:BaseDados}
\end{center}
\end{figure}

Figure \ref{fig:BaseDados} summarizes the changes made on the original dataset. Since the dataset is not labeled, we exploited techniques for semi-supervised learning. Due to the large size of the dataset, the labeling process was automated data following the logical criteria set by Bartle each of the four behaviors, and will be presented in Section IV.

\subsection{Used Technologies}

Data transformation operations were performed in Python and data analysis was performed using RapidMiner \cite{mierswa2006}. We choose RapidMiner as an analysis tool by several reasons: it is a consolidated package on the market (work started in 2001), it has a large installed base of users (more than half a million) and active communities for immediate support; it was acknowledged by KDnuggets \cite{Kdnug} as the best tool for data mining and in 2013; it has more than 500 operators for all professional analysis purposes; and, finally, it is presented as a tool with an easy learning curve.

\section{Experiments}

The dataset used in \cite{LEE2011} shows 91,065 players and 667,032 distinct sections of play, containing only data of the Horde. Further, the dataset did not contain data associated with exchange of messages between the players, which would help to identify the behaviors of Bartle. However, with the new processed dataset, as shown in Figure \ref{fig:BaseDados}, it was possible to obtain other useful information which is important to identify the behaviors, as explained in the previous section, such as ``Evolution'', ``Speed of evolution'' and others. Following the observations of Bartle, we traced the following guidelines for each behavior:

\small
\begin{itemize}

\item \emph{Killer}: high level, few maps, but maps to all high level or, in case of PvP, neutral level maps;

\item \emph{Socializer}: beginners, maps stagnated in neutral or newbie, when developments is presented, this is the basic level up (fast) and not more;

\item \emph{Achiever}: a lot of playing time and high levels of development;

\item \emph{Explorer}: a lot of playing time and visited many maps, not concerned with the evolution of levels.

\end{itemize}
\normalsize

Thus, the following rules have been created to the automation annotation labels' step in semi-supervised learning:

{\fontsize{4}{4}\selectfont$
\mathbf{Killer}: EndLevel \geq 60 \wedge (MapHigh + MapMiddle + MapNeutral > 70) \wedge NMaps < 10 \wedge LevelSpeed \leq 25
$}

{\fontsize{4}{4}\selectfont$
\mathbf{Socializer}: EndLevel \leq 30 \wedge (MapNovice + MapNeutral > 30) \wedge LevelSpeed < 15
$}

{\fontsize{4}{4}\selectfont$
\mathbf{Achiever}: Time \geq 1800 \wedge NMaps < 25 \wedge LevelSpeed \geq 25
$}

{\fontsize{4}{4}\selectfont$
\mathbf{Explorer}: Time \geq 1800 \wedge NMaps \geq 30 \wedge LevelSpeed < 25
$}

\begin{figure}[!t]
\begin{center}
\includegraphics[scale=0.9]{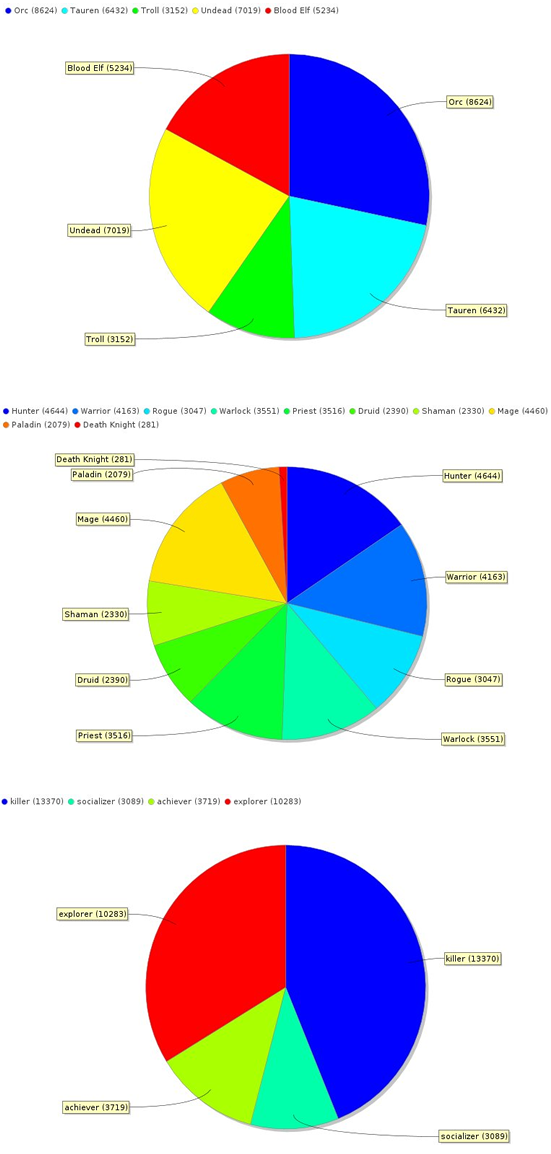}
\caption{Graphs showing the relationship between the number of players belonging to the Races and Classes and types of WoW players presented by Bartle.}
\label{fig:Estatisticas}
\end{center}
\end{figure}

Was taken as a basis for the values set in each equation the values considered by players as being high or low values. This step adjustment values was made with three regulars WoW players known to the authors. For the time regulation, it was used as a parameter of the research \cite{LIVESCIENCE2013}, where researchers show that 13 hours of play per week is considered high value --- in the case of this study, the total is computed for 3 hours years of data collection, and correspond approximately 1800 hours to 12,5 hours per week, or almost 2 hours per day.

Based on the previous equations we automatically labeled 30\% of the dataset, and the remainder of the dataset was labeled using a semi-supervised learning algorithm. For the tests, we performed the evaluation by decision tree, with 5-fold cross-validation, information gain as criteria, minimal leaf size equals 2, minimal gain of 0,1, maximal depth 20, confidence equals 0,01, and observed the results. While building the tree, we used features such as ``Class'', ``Initial Level'', ``Final Level'', ``Evolution'', ``Number of zones visited'', ``Race'', ``Game Time'' and ``Behavior''.

The tests were performed separately, considering the years 2006, 2007, 2008 and 2009 in isolation, and also a final test grouping all data from 2006 to 2009. Tables and graphs herein refer to tests from 2006 to 2009, and when necessary, a test is indicated isolated by specific year. This analysis taken in a year-basis is important as every year a new update patch was released, changing the dynamics of the game and thus enabling the analysis of the impacts of these changes on the game player behaviors.

Figure \ref{fig:Estatisticas} shows the number of players belonging to each of the races. It is worth noticing that the races presented are only related to Horde players, as previously mentioned. Clearly, the most popular races are Orcs and Undead, while there is an average preference for Blood Elves and Tauren, and a very low preference for Trolls. These preferences are due to the classes that are associated with each race. For instance, classes such as Hunters, Mages and Warrior are the most preferred ones, possibly because they are easier to play or just more funny. There is a pre-specified relationship between classes and races, and thus, some classes are only available for some races.

\begin{figure}[!t]
\begin{center}
\includegraphics[scale=0.4]{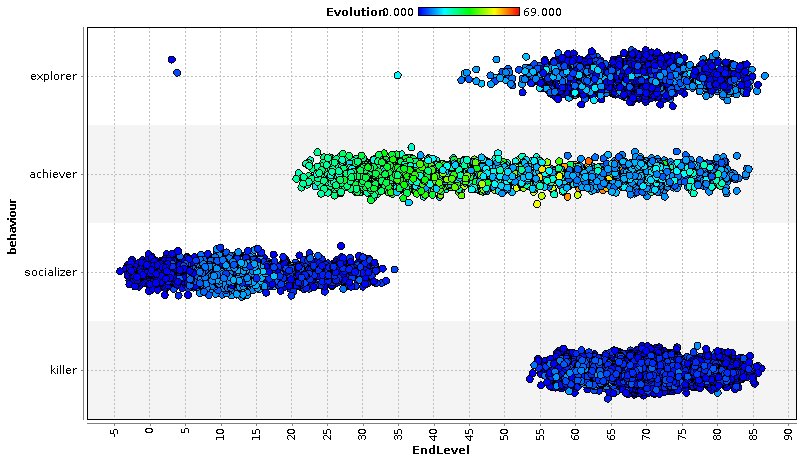}
\caption{Relationship between players' behaviors and their levels.}
\label{fig:Scatter-EndLevelxBehaviour-Color-Evolution}
\end{center}
\end{figure}

It is observed in the second graph in Figure \ref{fig:Estatisticas}, which shows a preference for classes, the Hunters have advantages, this is due to it owning a pet unique (standard monster auxiliary) that appeals to enough players and also because of their quick and easy level gain over other classes. Another class that is fairly chosen because of its natural pet is the Mage, which also has major damage in an area that facilitates large kill hordes of monsters quickly. When looking for a relationship between these preferences and player types proposed by Bartle, note that the classes that are most preferred are those that have great damage, easy/rapid evolution and are effective for use in PvP. The third graph shows that killers represent most players, which may be caused by the great preference for rapid, high-class damage. Hunters can also fit the characteristic of explorers. Because they have great mobility, field of creatures and ease of development, it can be inferred that possibly Hunters and Warriors snap heavily on behaviors and explorer killer. The other classes can fit into different behaviors, as well as those mentioned above can be used by Socializers and achievers, all depending on the skill and preference of a player gameplay.

\begin{figure}[!t]
\begin{center}
\includegraphics[scale=0.4]{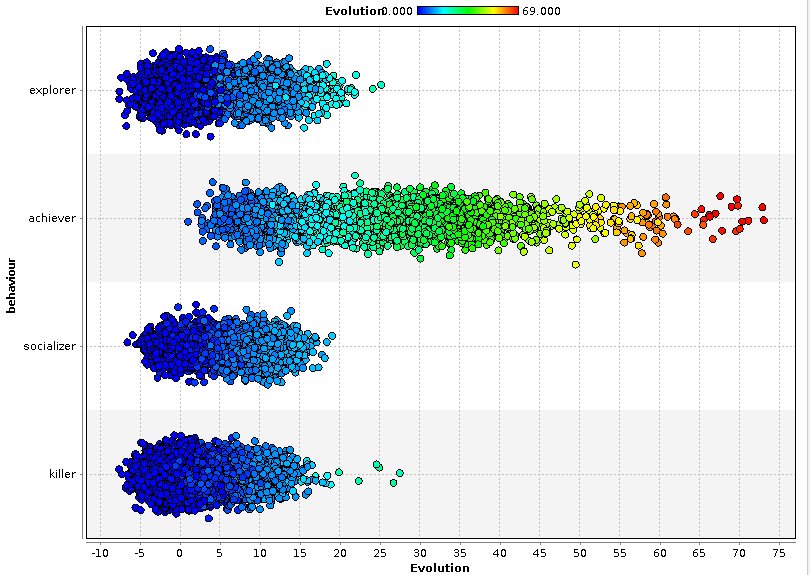}
\caption{Relationship between players' behaviors and their evolution.}
\label{fig:Scatter-EvolutionxBehaviour-Color-Evolution}
\end{center}
\end{figure}

\begin{figure}[!t]
\begin{center}
\includegraphics[scale=0.4]{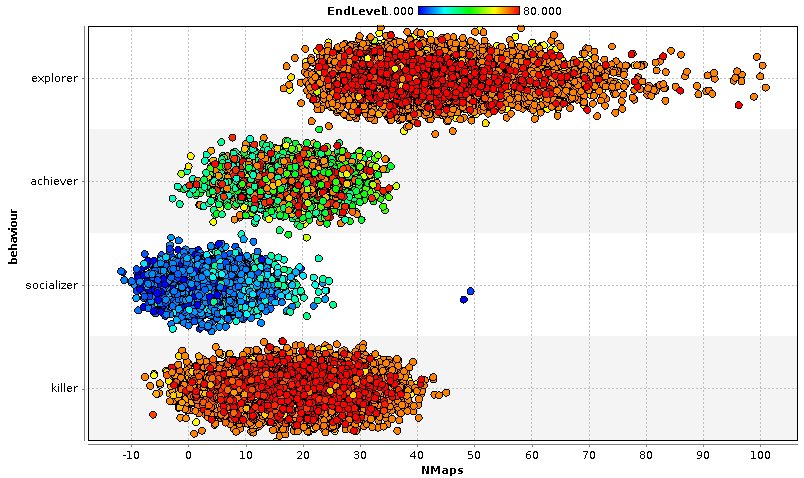}
\caption{Relationship between players' behaviors and the number of zones visited.}
\label{fig:Scatter-NMapsxbehaviour-colorEndLevel}
\end{center}
\end{figure}

It is observed in Figure \ref{fig:Scatter-EndLevelxBehaviour-Color-Evolution} as the development of the players behavior. It may be noted striking characteristics for each of the behaviors: the explorers tend to have a medium to slow progress, but we note that these are average to high, therefore, that they may walk freely in the world without being killed by monsters considered aggressive, it is necessary to have sufficient to annihilate these monsters. The achievers in turn have rapid development, as they try to develop your character doing quests and using the appropriate maps for training. It is then observed that the achiever usually visit more maps than the killers, for example, but far fewer maps that explorers. Note that the achievers are players with very high rate of evolution. The socializers usually have subdued to a certain level and when they reach a comfortable level, where they can use all the items considered fashions offered by the game, stop evolving. Since the main objective of this type of player is to socialize, gather props, pets and other fashions - so he does not need to continue evolving since its habitat are neutral cities and not the wild and PvP maps. The killers are usually high-level players who aim to actively participate in PvP, raids and camps (stand in certain places to attack other characters). This type of player usually attends some maps, but not as many as achievers and explorers, preferring maps very high level, because they can survive there, or to places of PvP in general cities. An interesting observation is that many players end up changing behavior when they reach certain characteristics. An achiever, for example can become a killer or an explorer, as when it reaches the highest levels of the quests can become scarce or even at the highest level. Therefore one possibility is that explorers and killers can be classes as arising from achievers. However, the dataset \cite{LEE2011} does not provide enough information to determine this type of migration between behaviors.

It is observed in Figure \ref{fig:Scatter-EvolutionxBehaviour-Color-Evolution} the relationship between the behaviors presented by Bartle and the evolution of the players. Note that the term evolution denotes the level variation, that is, the final level of the player minus the initial level during the period of data collection.

Observing this graph alone for behaviors explorer, socializer and killer, does it show a large discrepancy between the evolution. But an analysis contrasting Figure \ref{fig:Scatter-EvolutionxBehaviour-Color-Evolution} and Figure \ref{fig:Scatter-EndLevelxBehaviour-Color-Evolution} allows gamers to realize that stagnating at different levels. The socializers are at lower levels, while the killers and explorers fail to evolve at higher levels. When we analyze the achievers, we note that most of these have high rates of evolution and realize that this evolution finishes for making these players have a level of mid to high end. The speed of evolution of the same can be given by the large amount of experience won through quests in the game and also the efficient management of resources by the player, and the correct use of the world (to know where to play to evolve faster) to obtain quick experience.

It might also analyze the data from the perspective of the visited number of maps in order to make a better separation between two types of players who would rather diffuse, and the killer explorer. Note that the data analyzed so far, does not say a lot about the separation of these two types of players. The main feature of the explorer is to visit as many maps as possible, in order to know the world of games more accurately, as well as local secret and unexplored. The killer, meanwhile, can also visit many maps in order to recognize them and establish strategies for PvP, persecution and escape from enemies, but these players move from one map to the other with a frequency slightly smaller and not park the same (the maps are only access roads to places where they want to reach). It is observed in Figure \ref{fig:Scatter-NMapsxbehaviour-colorEndLevel} so that the amount of maps explorers is much higher than any other type of player ranked. The Socializers visit fewer maps, since they tend to park on maps where en-contros happen and they are safe from killers. Finally, the achievers have small displacement, since they can only focus on one area in order to accomplish all the quests in this area and evolve without major problems.

\begin{figure*}[!t]
\begin{center}
\includegraphics[scale=0.6]{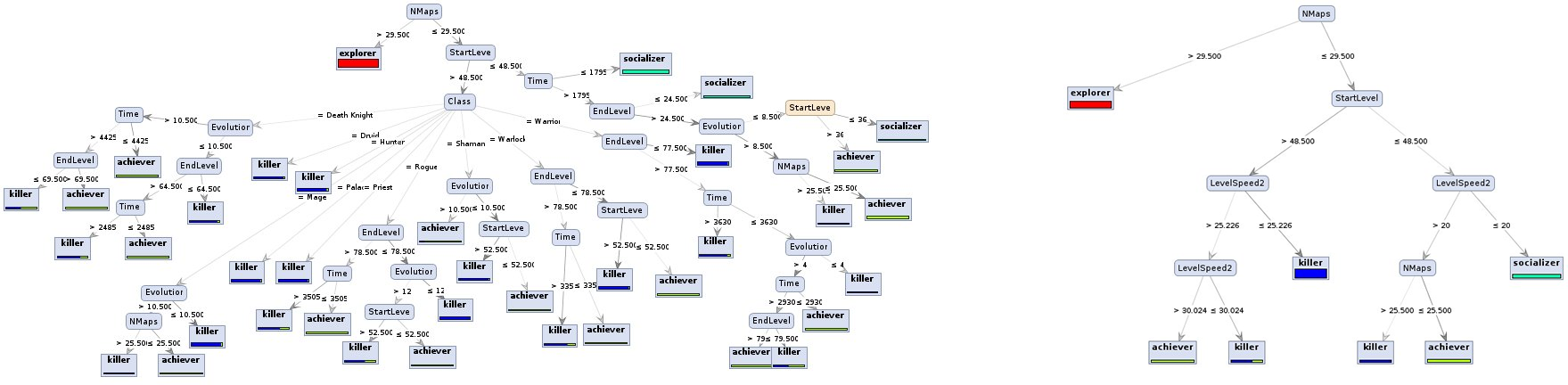}
\caption{Comparison between a decision tree generated with all possible attributes (left), except the `` Speed of evolution'', and one that considers this attribute (right), for the 2008 dataset.}
\label{fig:LevelSpeed}
\end{center}
\end{figure*}

It was also possible to observe something very interesting in the theory of ML, as shown in Figure \ref{fig:LevelSpeed}. A test was made of tree generation considering all the attributes of the base (left panel), except for the ``Speed of evolution'', and a test including the ``Speed of evolution'', keeping all other configuration parameters validation and tree. What has been observed is that the left tree was much larger and included in your solution attributes such as ``Initial Level'', ``Final Level'', ``Evolution'', ``Class'', ``Race'' and others. However, the tree on the right, which uses an attribute that condenses several of these, became smaller. It was also observed that the results were slightly better accuracy in the second case, and can realize that where possible the creation of an attribute that condense the information from various other attributes are not only reduces the complexity of the solution to improve its performance.

\begin{figure}[!t]
\begin{center}
\includegraphics[scale=0.5]{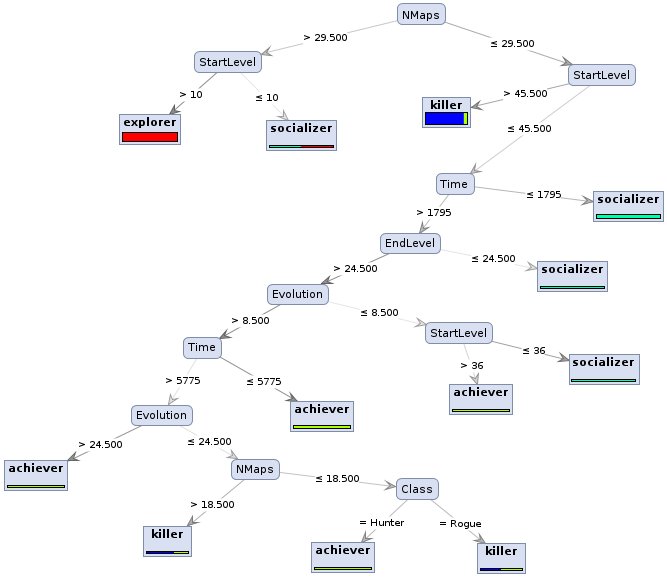}
\caption{Decision tree for the dataset collected from WoW, 2006-2009.}
\label{fig:Tree}
\end{center}
\end{figure}

As can be seen from Table \ref{tab:Resultados}, the accuracy achieved by the proposal of this work was very good. We credited to this fact a very clear difference between each player behavior proposed by Bartle, ie, there is always at least one criterion that distinguishes one behavior from another where they could be confused. Figure \ref{fig:Tree} illustrates the result of a decision tree obtained for a condensed dataset from 2006 to 2009 and it shows as the attributes of better information gain are the same ones who Bartle presented as predominant (and important) for each behavior. The simplest behavior to identify was the explorers, once its most outstanding feature is that they visit many maps and this is quite distinct in relation to other behaviors.

Furthermore, the results indicated by the third graph in Figure \ref{fig:Estatisticas} and Figure \ref{fig:Bartle2} clearly show the analysis made by Bartle: the most notable is the presence of killers that inhibit the presence of socializers, as conjectured.

It is worth mentioning that it was also possible to identify the Game Masters (GM) of the game. The GM are Blizzard employees that possess the function to remain in local of high concentration of players (such as cities and neutral maps) observing their behavior regarding compliance with the rules of the game, for example, do not use inappropriate words, not disrupt the play of others etc. Thus, these users act as the police of modern societies. They have the function to mute the player, preventing they to speak publicly or in groups, and also, in more serious cases of infringement, ban the user from the game. The GMs have 0 evolution, remaining at level 1 for many hours connected (in one case, the GM was logged for 18 hours a day during the 3 years).

\section{Conclusion}

The work presented here was to develop a methodology to classify the players' behaviors proposed by \cite{BartleArtigo} in the game \emph{World of Warcraft} (WoW) and then evaluate the impact of the relationship between behaviors for that game. For this, we used the dataset collected by \cite{LEE2011} from 2006 to 2009 and we transformed that original dataset as a new one more useful for the task proposed.

One of the most important applications of the \emph{player modeling}'s identification is to help game developers or game designers to create experiences with appeals specifically devoted to each behavior: new areas/cities for socializers to chat and play, harder enemies for killers, maps more complete/complex for explorers and advanced class system for achievers are solutions that can improve the interactivity and gameplay of players.

Therefore, it was possible to automatically label part of the original dataset with each of the four behaviors proposed by Bartle: \emph{killer}, \emph{achiever}, \emph{socializer} and \emph{explorer}. The result of that was a dataset with a few labeled data, so we used the technique of semi-supervised learning, ideal for this type of situation. Labeling of the data followed the criteria observed by Bartle and parameter setting was done with the assistance of players of WoW.

It was observed that there is in WoW a preference for certain types of classes and races, but one of the behaviors excelled more than others: the killers. This may be due to the fact that WoW is an online game where there are several characteristics identified by Bartle that favor its appearance: large maps, plenty of weapons, many monsters to fight, and the main one, PvP areas liberated in any place (through challenge agreed by both players) and specific areas for PvP (in the cities). For the same reasons, except for the PvP areas, there is a preference for the explorer: WoW is a game with 229 immense maps, rich in details (tridimensional modeling, art, equipaments, quests and so on) and stories that intersect, thus making it an ideal game for that kind of players.

\begin{table*}
\center
\begin{tabular}{c|c|c|c|c|c|}
\cline{2-6}
  & \medio 2006 & \medio 2007 & \medio 2008 & \medio 2009 & \medio 2006 a 2009             \\ \cline{2-6}
 \multicolumn{1}{ |c| }{\medio \emph{Killers}}          & 97.62 / 98.97 & 88.27 / 99.94 & 98.07 / 99.96 & 98.90 / 98.90 & 92.45 / 99.96          \\ \cline{1-6}
 \multicolumn{1}{ |c| }{\medio \emph{Socializers}}      & 100 / 99.66   & 100 / 99.71   & 100 / 100     & 95.65 / 100   & 99.94 / 99.94          \\ \cline{1-6}
 \multicolumn{1}{ |c| }{\medio \emph{Achievers}}        & 99.58 / 99.03 & 99.38 / 58.93 & 99.75 / 89.43 & 67.86 / 65.52 & 99.81 / 70.66          \\ \cline{1-6}
 \multicolumn{1}{ |c| }{\medio \emph{Explorers}}        & 99.81 / 100   & 100 / 100     & 100 / 100     & 100 / 100     & 99.98 / 99.98          \\ \cline{1-6}
 \\ \cline{1-6}
 \multicolumn{1}{ |c| }{\medio Accuracy}                & 99.55 $\pm$ 0.09 & 94.46 $\pm$ 0.33 & 99.01 $\pm$ 0.12 & 98.01 $\pm$ 0.70 & 96.39 $\pm$ 0.27       \\ \cline{1-6}
\end{tabular}
\endcenter
\caption{Results of the experiments performed. Each value for the bahaviors indicates its precision/recall. The last row shows the accuracy of each experiment.}
\label{tab:Resultados}
\end{table*}

For some cases observed in the tests, only the data collected would not be sufficient to classify the players between behaviors exploited. As suggested by Bartle, it is important to add the written expression of the players (through the collection of words in chats) as this would assist in determining a behavior where there was doubt in the player rankings.

The technique presented here has been shown quite satisfactory to define the behaviors of the players in WoW. But it is important to note that the data used does not allow us to verify, for example, if a high level player, who previously was an achiever (by evolving rapidly), became a socializer. This analysis would only be possible by monitoring, for example, logs of conversations or individual player behaviour. Thus, migration from one type to another has not been included in this work and becomes an incentive for analysis in future work. Other interesting works as expansion of this one may include chat analysis (capture a new base) to improve the classification as already discussed, identifying subtypes player (80\% killer and 20\% explorer, for example) and explore the fusion of attributes as analyzed in Figure \ref{fig:LevelSpeed}, since the results presented here were very interesting (reducing the size of the decision tree and accuracy improvement).

\bibliographystyle{IEEEtran}
\bibliography{Artigo}

\end{document}